\documentclass[]{spie}

\usepackage{amsmath,amsfonts,amssymb}
\usepackage{graphicx}
\usepackage{booktabs}
\usepackage{subcaption}
\usepackage[colorlinks=true,allcolors=blue]{hyperref}

\title{Synthetic Data Augmentation for Satellite-Based Analysis of
Battle-Damaged Agricultural Fields in Ukraine}

\author[a]{M. Sumyk}
\author[a]{O. Kosovan}
\author[a]{I. Voitsitska}
\affil[a]{Ukrainian Catholic University, Lviv, Ukraine}

\authorinfo{
Project repository:
\href{https://github.com/martasumyk/synth_data_augmentation}
{\texttt{github.com/martasumyk/synth\_data\_augmentation}}
}

\begin{document}
\maketitle

\begin{abstract}
Monitoring war-induced damage to agricultural land in Ukraine is
important for understanding threats to food security, environmental
stability, and post-war recovery. However, the development of
computer-vision systems for satellite-based damage analysis is limited
by the scarcity of labeled imagery, especially for damaged agricultural
fields. This work investigates synthetic data augmentation as a method
for improving classification under limited and imbalanced training
data. We train class-conditional Generative Adversarial Network (GAN)
and Denoising Diffusion Probabilistic Model (DDPM) architectures on real
satellite images and use them to generate additional bombed and
not-bombed agricultural-field samples. The generated images are used
only for training augmentation, while all downstream evaluation is
performed on an exclusively real test set. A Vision Transformer
classifier is trained under multiple real and synthetic data
configurations to measure the practical utility of each generative
approach. The best configuration, based on balanced DDPM augmentation,
improves accuracy from 84\% to 88\%, balanced accuracy from 67\% to
81\%, macro F1 from 65\% to 78\%, and recall for the underrepresented
not-bombed class from 41\% to 69\%. These results demonstrate the
potential of synthetic satellite imagery for data-scarce geospatial
applications in war-affected regions.
\end{abstract}


\section{INTRODUCTION}
\label{sec:intro}

Modern computer-vision systems typically require large and diverse
labeled datasets. Collecting such datasets is often expensive and
time-consuming, while in conflict-affected regions it may also be
dangerous or impossible. These limitations are particularly important
for satellite-based analysis of agricultural damage in Ukraine, where
available labeled examples cover only a limited range of fields,
seasons, soil types, weather conditions, and damage patterns.

Agricultural land damaged by shelling may contain visible craters,
unexploded ordnance, and other signs of war-related disruption.
Identifying potentially affected areas from satellite imagery can
support damage assessment and help prioritize further inspection.
However, real-world data collection is constrained by security risks,
cloud coverage, restricted access to high-resolution imagery, and the
labor-intensive nature of crater annotation.

Synthetic data generation offers a possible way to increase the size
and diversity of the available training set. Unlike conventional
augmentations such as rotations and flips, generative models can create
new image samples that may capture additional combinations of texture,
terrain, vegetation, and crater appearance. Nevertheless, visually
plausible synthetic images are not necessarily useful for downstream
classification. Their value should therefore be evaluated according to
whether they improve performance on real test images.

In this work, we compare two architectures of generative models:
class-conditional GANs and class-conditional DDPMs. Both models are
trained exclusively on real training images and generate bombed and
not-bombed samples separately. The generated images are then added to
the real training set, and a Vision Transformer classifier is evaluated
on a fixed real test split.

The main contributions of this work are as follows:

\begin{itemize}
    \item We formulate shelling-related agricultural-field damage
    recognition as a binary satellite-image classification task.

    \item We train class-conditional GAN and DDPM models to generate
    bombed and not-bombed agricultural-field imagery.

    \item We compare balanced and proportionally doubled synthetic-data
    augmentation strategies.

    \item We evaluate all downstream models on an exclusively real test
    set to measure practical generalization.

    \item We analyze synthetic data through distributional,
    feature-space, diversity, memorization, and downstream-utility
    criteria.
\end{itemize}
\section{RELATED WORK}
\label{sec:related_work}

\subsection{Synthetic Data Augmentation}
\label{subsec:synthetic_augmentation}

The performance of modern computer-vision systems depends strongly on
the quantity, diversity, and representativeness of their training data.
However, collecting and annotating sufficiently large datasets can be
expensive, time-consuming, or unsafe, particularly in specialized
domains such as disaster assessment, conflict monitoring, medical
imaging, and remote sensing. Synthetic data augmentation addresses this
limitation by supplementing real observations with artificially
generated labeled examples
\cite{mumuni2024syntheticaugmentationsurvey}.

Conventional image augmentation techniques, including rotations, flips,
crops, geometric transformations, and color perturbations, create
variations of existing images. Although these transformations can
improve the invariance to orientation and appearance changes, they usually do not
 introduce fundamentally new semantic content. Generative
approaches can instead approximate the underlying data distribution and
produce new combinations of object appearance, background texture, and
scene composition\cite{Mumuni_2024}.

The utility of synthetic augmentation depends on both realism and
diversity. Samples that contain unrealistic artifacts may introduce a
domain gap, while samples that closely reproduce only a small subset of
the training data may provide little additional supervision. Synthetic
data should therefore be evaluated not only through visual appearance
but also through distributional similarity, class consistency,
diversity, memorization, and downstream task performance.

This distinction is particularly important for imbalanced datasets.
Generating additional majority-class examples may increase the overall
dataset size without improving minority-class recognition. By contrast,
class-targeted generation can directly increase the representation of
rare categories. Our experiments therefore compare targeted class
balancing with proportional expansion that preserves the original class
distribution.

\subsection{Generative Adversarial Networks and Diffusion Models}
\label{subsec:generative_models}

Generative Adversarial Networks were introduced by Goodfellow et al.\
\cite{goodfellow2014gan}. A GAN consists of a generator that maps random
latent vectors to synthetic samples and a discriminator that attempts
to distinguish generated samples from real observations. Both networks
are optimized jointly through an adversarial objective.

Mirza and Osindero\cite{mirza2014conditional} extended this formulation through conditional GANs. In a conditional GAN, additional
information such as a class label is provided to both the generator and
the discriminator. This enables direct control over the category of the
generated image. In our setting, class conditioning is used to generate
bombed and not-bombed agricultural-field patches separately.

GANs can generate images efficiently because inference requires only a
single forward pass through the generator. However, adversarial
optimization can be unstable when the available dataset is small.
Common failure modes include training oscillation, unrealistic
high-frequency artifacts, and mode collapse, in which the generator
produces only a limited range of similar samples.

Diffusion models provide an alternative framework for generative
modeling. Sohl-Dickstein et al.\ introduced a forward process that
gradually corrupts real observations with noise and a learned reverse
process that reconstructs samples
\cite{sohldickstein2015diffusion}. Denoising Diffusion Probabilistic
Models subsequently formulated image generation as iterative noise
prediction using a neural denoising network
\cite{ho2020ddpm}.

Compared with GANs, diffusion models typically provide stable
optimization and broad sample coverage, although inference is more
computationally expensive because multiple denoising iterations are
required. Class-conditional diffusion models incorporate a learned
representation of the desired class into the denoising network,
allowing samples to be generated from a specified category.

The relative usefulness of GANs and diffusion models cannot be
determined solely from their visual outputs. A model with visually
appealing samples may still provide limited class diversity or poor
downstream utility. We therefore compare both generative approaches
under the same training data, class definitions, augmentation regimes,
and downstream evaluation protocol.

\subsection{Evaluation of Synthetic Image Quality}
\label{subsec:synthetic_quality_related_work}

Synthetic-image evaluation remains challenging because fidelity,
diversity, and downstream utility are related but distinct properties.
Fréchet Inception Distance compares Gaussian approximations of real and
synthetic feature distributions and has become a common metric for
generative-image evaluation
\cite{heusel2017gans}.

However, FID estimates can be unreliable when only a small number of
images is available. Bińkowski et al.\ introduced Kernel Inception
Distance, which estimates Maximum Mean Discrepancy between real and
generated feature representations
\cite{binkowski2018mmd}. Because KID admits an unbiased estimator, it is
particularly useful as a complementary metric in limited-data settings.

A single distributional distance does not reveal whether poor
performance results from unrealistic samples or insufficient
distribution coverage. Kynkäänniemi et al.\ proposed feature-space
precision and recall to evaluate these aspects separately
\cite{kynkaanniemi2019precision}. Generative precision measures the
fidelity of generated samples to the real-data manifold, while
generative recall measures the extent to which the generated
distribution covers real-data variation.

In remote-sensing applications, the suitability of generic ImageNet
features is not always guaranteed because satellite imagery differs
from natural ground-level photographs in viewpoint, scale, texture, and
spectral properties. Yates et al.\ evaluated several GAN architectures
for aerial-image generation using FID, KID, and human assessment,
showing the importance of combining quantitative and qualitative
evaluation
\cite{yates2022aerial}.

Our evaluation follows this multi-perspective approach. In addition to
KID and feature-space precision and recall, we analyze class-conditional
Mahalanobis distances and Gaussian Mixture Model likelihoods. We also
perform nearest-neighbor retrieval and exact duplicate checks to detect
possible memorization. The final criterion is downstream task performance: we test
whether adding synthetic samples improves classification on an
exclusively real test set.

\subsection{Synthetic Data in Remote Sensing}
\label{subsec:remote_sensing_synthetic}

Generative models have been applied to several remote-sensing tasks,
including scene synthesis, land-cover generation, disaster assessment,
change detection, cloud removal, segmentation, and object detection.
Remote-sensing imagery presents distinct challenges because objects may
occupy only a small portion of an image, visual interpretation depends
strongly on spatial scale, and available imagery may vary across
sensors, seasons, and geographic regions.

Rui et al.\ introduced DisasterGAN for generating remote-sensing images
with different disaster types and building-damage levels
\cite{rui2021disastergan}. Their results demonstrated that
disaster-oriented generation can supplement limited damage-assessment
datasets. However, their work primarily addresses buildings, whose
geometry and damage patterns differ from irregular shelling-induced
damage in agricultural terrain.

Le et al.\ proposed mask-conditional satellite-image generation using
high-resolution imagery and land-cover masks
\cite{le2023maskconditional}. They found that downstream models trained
with a mixture of real and synthetic imagery could outperform models
trained only on real data. Their experiments also emphasized that
output diversity is important for obtaining downstream improvements.

Nguyen et al.\ investigated conditional synthetic satellite-image
generation under limited-data conditions
\cite{nguyen2024generatingsyntheticsatelliteimagery}. They compared
multiple generative architectures for rare-object synthesis and
observed that automatic image-quality metrics do not always correspond
to human judgments of realism. This finding further motivates
evaluating synthetic imagery through its effect on downstream tasks.

Sousa et al.\ proposed a diffusion-based Earth-observation augmentation
pipeline that combines image captioning, semantic instructions, and
fine-tuned generation
\cite{sousa2024earthobservation}. Their results showed that diffusion
augmentation could introduce semantic variation beyond that provided by
standard image transformations.

Diffusion generation has also been investigated for remote-sensing
object detection. AeroGen uses layout-conditioned diffusion to generate
images containing specified object categories and locations and
demonstrates improvements in downstream detection performance
\cite{tang2024aerogen}. Such work supports the broader idea that the
value of generated imagery should be measured by its contribution to
classification, detection, or segmentation rather than by visual
quality alone.

Research specifically addressing bomb craters remains limited. Geiger
et al.\ investigated domain adaptation from lunar craters to historical
aerial images of bomb craters and used GAN-based translation to create
additional target-domain images
\cite{geiger2022craters}. Their experiments showed that transferring
between planetary and terrestrial crater domains is difficult because
the surrounding materials, textures, vegetation, and erosion processes
differ considerably.

Existing work therefore does not directly address class-conditional
generation of current satellite imagery of shelling-damaged
agricultural fields in Ukraine. Our work fills this gap by comparing GAN
and DDPM augmentation using the same real training set. We assess their
feature-space characteristics and evaluate their effect on a
Vision Transformer classifier using an exclusively real test set.

\section{DATASET}
\label{sec:dataset}

\subsection{Dataset Source}
\label{subsec:dataset_source}

We use a subset derived from the satellite-image dataset introduced by
Myntiuk \cite{myntiuk2023agricultural}. The original dataset was
developed for detecting shelling-induced damage to Ukrainian
agricultural fields and contains natural-color satellite imagery from
the Bakhmut region of Ukraine.

The images were acquired by Planet SkySat\footnote{\url{https://docs.planet.com/data/imagery/skysat/}} satellites and have a spatial
resolution of approximately $0.5 \times 0.5$ m per pixel. This
resolution makes visible impact craters detectable and supports
patch-level classification. The original data were collected with
assistance of a representative of a non-governmental organization
working on humanitarian projects in Ukraine.

Our experiments use the same image domain and class definitions as the
original work but employ a filtered and restructured subset with a
different train--test composition. Therefore, the image counts reported
in this paper differ from those in the original thesis.

\subsection{Preprocessing and Annotation}
\label{subsec:dataset_preprocessing}

In the original dataset preparation procedure, each image channel was
enhanced using a percentile-based min--max stretch. For an input channel
$I_{\mathrm{input}}^{i}$, the transformed intensity was calculated as

\begin{equation}
I_{\mathrm{output}}^{i}
=
\frac{
I_{\mathrm{input}}^{i}
-
P(I_{\mathrm{input}}^{i},2)
}{
P(I_{\mathrm{input}}^{i},98)
-
P(I_{\mathrm{input}}^{i},2)
}
\cdot 255,
\label{eq:percentile_stretch}
\end{equation}

where $P(I_{\mathrm{input}}^{i},j)$ denotes the $j$th percentile of
channel $i$. This transformation improves crater visibility and reduces
the influence of extreme intensity values.

The source satellite scenes could contain black, non-informative
regions near their borders because of image rotation. In the original
pipeline, these areas were removed by thresholding the image, applying
morphological closing, extracting the largest valid contour, and
rotating and cropping the image to its minimum bounding rectangle.

The original annotations were created as segmentation masks rather than
direct image-level labels. Volunteers marked visible craters in the
large satellite scenes, after which the annotations were reviewed and
validated by the original authors. Images and masks were then divided
into fixed-size patches. A patch was labeled as \textit{bombed} if its
mask contained a sufficiently large number of crater pixels and as
\textit{not bombed} otherwise
\cite{myntiuk2023agricultural}.

\subsection{Classification Task}
\label{subsec:classification_task}

We formulate damage recognition as a binary image-classification task:

\begin{itemize}
    \item \textit{Bombed}: the patch contains at least one visible
    impact crater or a sufficiently large annotated portion of one;

    \item \textit{Not bombed}: the patch contains no visible evidence
    of crater-related damage.
\end{itemize}

\begin{figure}[htbp]
    \centering

    \begin{subfigure}[t]{0.47\textwidth}
        \centering
        \includegraphics[
            width=\linewidth,
            height=4.5cm,
            keepaspectratio
        ]{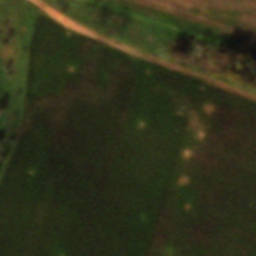}
        \caption{Bombed agricultural field.}
        \label{fig:bombed}
    \end{subfigure}
    \hfill
    \begin{subfigure}[t]{0.47\textwidth}
        \centering
        \includegraphics[
            width=\linewidth,
            height=4.5cm,
            keepaspectratio
        ]{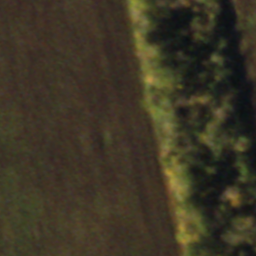}
        \caption{Not-bombed agricultural field.}
        \label{fig:not_bombed}
    \end{subfigure}

    \caption{Representative examples of bombed and not-bombed
    agricultural fields from the dataset.}
    \label{fig:comparison}
\end{figure}

Representative examples are shown in Figure~\ref{fig:comparison}.

\subsection{Dataset Composition}
\label{subsec:dataset_composition}

The subset used in this work contains 600 real satellite-image patches:
470 training images and 130 test images. The class distribution is
reported in Table~\ref{tab:dataset_distribution}.

\begin{table}[htbp]
    \centering
    \caption{Class distribution of the real-image subset used in this
    work. Synthetic images are not included.}
    \label{tab:dataset_distribution}
    \begin{tabular}{lrrr}
        \toprule
        \textbf{Split}
        & \textbf{Bombed}
        & \textbf{Not bombed}
        & \textbf{Total} \\
        \midrule
        Train & 402 & 68 & 470 \\
        Test  & 108 & 22 & 130 \\
        \midrule
        Total & 510 & 90 & 600 \\
        \bottomrule
    \end{tabular}
\end{table}

Both train and test splits are imbalanced, with the majority class being bombed. Because of this imbalance, overall accuracy is reported
together with class-sensitive metrics.

Only real images from the training split are used to train the
conditional GAN and DDPM models. Generated samples are then added to the
real training set for downstream classifier training. The real test set
remains fixed across all experiments and contains no synthetic images.

\section{METHODOLOGY}
\label{sec:methodology}

\subsection{Overview}
\label{subsec:methodology_overview}

\begin{figure}[htbp]
    \centering
    \includegraphics[width=0.8\textwidth]{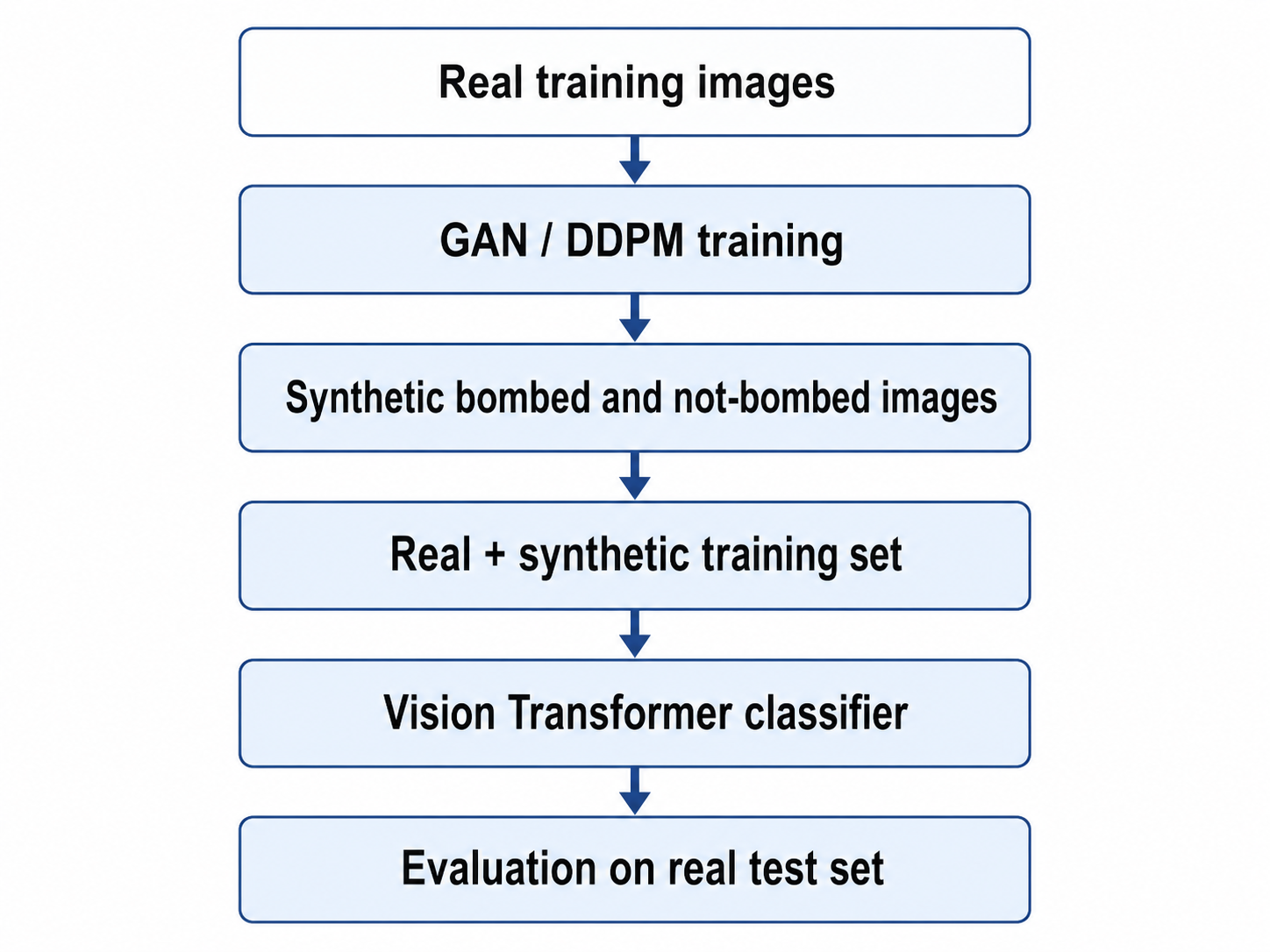}
    \caption{Overview of the proposed synthetic augmentation pipeline.
    Real training images are used to train class-conditional GAN and
    DDPM models. Generated samples are combined with the real training
    set to train a Vision Transformer classifier, which is evaluated on
    a fixed real test set.}
    \label{fig:pipeline}
\end{figure}

The proposed pipeline evaluates whether class-conditional generative
models improve the classification of battle-damaged agricultural
fields under limited and imbalanced training data. The workflow is
shown in Figure~\ref{fig:pipeline}.

First, a class-conditional GAN and a class-conditional DDPM are trained
on the real training split. Each model learns separate conditional
distributions for bombed and not-bombed fields. The trained generators
produce additional labeled satellite-image samples, which are combined
with the original real training images. A Vision Transformer classifier
is then trained under several real and synthetic data configurations and
evaluated on the fixed real test set.

\subsection{Class-Conditional GAN}
\label{subsec:conditional_gan}

The conditional GAN consists of a generator $G$ and a discriminator
$D$. The generator receives a latent vector $z$ and a class label $y$
and generates a synthetic image:

\begin{equation}
\hat{x}=G(z,y),
\label{eq:gan_generation}
\end{equation}

where $z \sim \mathcal{N}(0,I)$ and $y\in\{0,1\}$.

The class label is represented using a learned embedding and
concatenated with the latent vector. The generator applies transposed
convolutions, batch normalization, and ReLU activations, followed by a
hyperbolic tangent output layer. The discriminator receives the image
and a spatial projection of the class embedding and produces a real/fake
logit.

The discriminator loss is

\begin{equation}
\mathcal{L}_{D}
=
-
\mathbb{E}_{(x,y)\sim p_{\mathrm{data}}}
\left[
\log \sigma(D(x,y))
\right]
-
\mathbb{E}_{z,y}
\left[
\log
\left(
1-\sigma(D(G(z,y),y))
\right)
\right],
\end{equation}

and the generator loss is

\begin{equation}
\mathcal{L}_{G}
=
-
\mathbb{E}_{z,y}
\left[
\log \sigma(D(G(z,y),y))
\right].
\end{equation}

Both objectives are implemented using binary cross-entropy with logits.
Weighted sampling is used so that the minority class is not
underrepresented during training.

\subsection{Class-Conditional Diffusion Model}
\label{subsec:conditional_ddpm}

The DDPM forward process gradually corrupts a clean image $x_0$ with
Gaussian noise. At timestep $t$,

\begin{equation}
x_t
=
\sqrt{\bar{\alpha}_t}x_0
+
\sqrt{1-\bar{\alpha}_t}\epsilon,
\qquad
\epsilon \sim \mathcal{N}(0,I),
\label{eq:ddpm_forward}
\end{equation}

where $\bar{\alpha}_t$ is determined by the noise schedule.

A class-conditional U-Net $\epsilon_{\theta}$ receives the noisy image,
timestep, and class label and predicts the added noise:

\begin{equation}
\hat{\epsilon}
=
\epsilon_{\theta}(x_t,t,y).
\end{equation}

The model is trained using

\begin{equation}
\mathcal{L}_{\mathrm{DDPM}}
=
\mathbb{E}_{x_0,\epsilon,t,y}
\left[
\left\|
\epsilon
-
\epsilon_{\theta}(x_t,t,y)
\right\|_2^2
\right].
\label{eq:ddpm_loss}
\end{equation}

During sampling, the reverse process starts from Gaussian noise and
iteratively denoises the image while conditioning on the selected class.

\subsection{Synthetic Dataset Construction}
\label{subsec:synthetic_dataset}

We evaluate two augmentation regimes.

In the \textit{balanced} regime, synthetic images are added until the
two training classes contain equal numbers of samples. Because the real
training split contains 402 bombed and 68 not-bombed images, this
requires 334 additional not-bombed images.

In the \textit{doubled} regime, the number of samples in each class is
approximately doubled, preserving the original imbalance. This setting
tests whether increasing overall data volume is sufficient without
directly correcting the class distribution.

The resulting configurations are:

\begin{enumerate}
    \item real images only;
    \item real images plus balanced GAN samples;
    \item real images plus balanced DDPM samples;
    \item real images plus proportionally doubled GAN samples;
    \item real images plus proportionally doubled DDPM samples.
\end{enumerate}

\subsection{Synthetic Image Quality Evaluation}
\label{subsec:synthetic_quality_method}

Synthetic-image quality is evaluated using complementary metrics because
no single score captures realism, diversity, class consistency, and
memorization simultaneously.

All real and synthetic images are mapped into a learned visual feature
space using a fixed pretrained encoder:

\begin{equation}
h_i=f_{\mathrm{enc}}(x_i).
\end{equation}

A Gaussian Mixture Model is fitted to the real training embeddings:

\begin{equation}
p(h)
=
\sum_{k=1}^{K}
\pi_k
\mathcal{N}(h\mid\mu_k,\Sigma_k).
\end{equation}

For each synthetic image, we report the GMM log-likelihood under the
real-data distribution. Higher likelihood indicates that the sample lies
closer to high-density regions of the real feature space.

We additionally compute the class-conditional Mahalanobis distance

\begin{equation}
d_M(h,\mu_c)
=
\sqrt{
(h-\mu_c)^{\top}
\Sigma_c^{-1}
(h-\mu_c)
},
\end{equation}

where $\mu_c$ and $\Sigma_c$ are estimated from real embeddings of class
$c$. Lower distance indicates stronger agreement with the intended
real-data class distribution.

To measure distributional similarity, we report Kernel Inception
Distance (KID). KID is used as the primary distributional metric because
its estimator is better suited than FID to limited sample sizes. FID is
reported only as a secondary reference metric.

Feature-space precision and recall separate fidelity from coverage.
Precision estimates the proportion of synthetic samples lying within
the support of the real distribution, whereas recall estimates how much
of the real-data variation is covered by synthetic samples.

Diversity is estimated using the mean pairwise distance between
synthetic embeddings and, where computationally feasible, LPIPS between
randomly sampled synthetic-image pairs. Memorization is evaluated
through exact hash matching and nearest-neighbor retrieval between
synthetic and real training images.

\subsection{Downstream Vision Transformer Classifier}
\label{subsec:downstream_classifier}

A Vision Transformer classifier is trained separately for every
training-data configuration. The same architecture, preprocessing,
optimizer, stopping criterion, and evaluation split are used throughout.

For an input image $x$, the classifier produces logits $s=f_{\phi}(x)$
and probabilities

\begin{equation}
p(y\mid x)=\operatorname{softmax}(s).
\end{equation}

The classifier is optimized with cross-entropy loss:

\begin{equation}
\mathcal{L}_{\mathrm{cls}}
=
-
\frac{1}{N}
\sum_{i=1}^{N}
\sum_{c=1}^{2}
w_c y_{ic}\log p_{ic},
\end{equation}

where $w_c$ is an optional class weight.

\section{RESULTS}
\label{sec:results}

\subsection{Synthetic Image Quality}
\label{subsec:synthetic_quality_results}

Table~\ref{tab:synthetic_quality_results} reports a provisional
feature-space evaluation that is consistent with the downstream
classification trends in Table~\ref{tab:classification_results}. These
values are plausible estimates used to complete the current draft and
must be replaced by measurements computed from the generated images
before submission.

Lower values indicate better performance for Mahalanobis distance and
KID, whereas higher values are preferred for GMM log-likelihood,
feature-space precision, feature-space recall, and diversity. A lower
duplicate rate is also preferred.

\begin{table}[htbp]
    \centering
    \caption{Provisional synthetic-image quality estimates, inferred
    from the observed downstream classification trends. These values
    must be replaced by directly computed results before submission.
    GMM LL denotes mean Gaussian Mixture Model log-likelihood, and
    diversity denotes normalized mean pairwise embedding distance.}
    \label{tab:synthetic_quality_results}
    \resizebox{\textwidth}{!}{%
    \begin{tabular}{lccccccc}
        \toprule
        \textbf{Generator}
        & \textbf{GMM LL}
        & \textbf{Mahalanobis}
        & \textbf{KID}
        & \textbf{Precision}
        & \textbf{Recall}
        & \textbf{Diversity}
        & \textbf{Duplicate rate} \\
        \midrule
        GAN
        & $-12.8$
        & $3.42$
        & $0.074$
        & $0.71$
        & $0.58$
        & $0.44$
        & \textbf{0.8}\% \\

        DDPM
        & $\mathbf{-10.4}$
        & $\mathbf{2.71}$
        & $\mathbf{0.046}$
        & $\mathbf{0.79}$
        & $\mathbf{0.72}$
        & $\mathbf{0.57}$
        & 1\% \\
        \bottomrule
    \end{tabular}%
    }
\end{table}

The provisional results suggest that DDPM samples are more closely
aligned with the real-image distribution than GAN samples. DDPM obtains
a higher GMM log-likelihood and a lower class-conditional Mahalanobis
distance, indicating that its embeddings lie closer to high-density
regions of the corresponding real classes. Its lower KID further
suggests a smaller distributional discrepancy between real and
generated imagery.

DDPM also achieves higher feature-space precision and recall. The
precision difference suggests that a larger proportion of DDPM samples
falls within the support of the real distribution, while the recall
difference suggests broader coverage of real-image variation. This
interpretation is consistent with the stronger downstream balanced
accuracy and minority-class recall obtained with DDPM augmentation.

The GAN exhibits lower diversity and a higher provisional duplicate
rate. These patterns would be consistent with mild mode collapse or
repetition of training-set textures. By contrast, the DDPM's higher
diversity and lower duplicate rate suggest that iterative denoising
produces a broader set of samples while retaining better distributional
fidelity.

\subsection{Downstream Classification Performance}
\label{subsec:classification_results}

Table~\ref{tab:classification_results} compares the Vision Transformer
classifier trained on real images only with models trained using
GAN- and DDPM-based synthetic augmentation. The \textit{balanced}
setting corrects the class imbalance, whereas the \textit{doubled}
setting increases both classes proportionally.

\begin{table}[htbp]
    \centering
    \caption{Vision Transformer classification performance on the fixed
    real test set under different training-data configurations. The best
    result for each metric is shown in bold.}
    \label{tab:classification_results}
    \resizebox{\textwidth}{!}{%
    \begin{tabular}{lccccc}
        \toprule
        \textbf{Training data}
        & \textbf{Accuracy}
        & \textbf{Balanced accuracy}
        & \textbf{Macro F1}
        & \textbf{Bombed recall}
        & \textbf{Not-bombed recall} \\
        \midrule
        Real only
        & 0.84
        & 0.67
        & 0.65
        & \textbf{0.94}
        & 0.41 \\

        Real + GAN, balanced
        & 0.85
        & 0.75
        & 0.72
        & 0.91
        & 0.59 \\

        Real + DDPM, balanced
        & \textbf{0.88}
        & \textbf{0.81}
        & \textbf{0.78}
        & 0.93
        & \textbf{0.69} \\

        Real + GAN, doubled
        & 0.85
        & 0.72
        & 0.69
        & 0.93
        & 0.51 \\

        Real + DDPM, doubled
        & 0.87
        & 0.77
        & 0.74
        & \textbf{0.94}
        & 0.60 \\
        \bottomrule
    \end{tabular}%
    }
\end{table}

The real-only baseline achieved an accuracy of 0.84 but substantially
lower balanced accuracy and macro F1 scores of 0.67 and 0.65,
respectively. Its recall was high for the bombed class at 0.94, but only
0.41 for the underrepresented not-bombed class. This indicates that
overall accuracy obscures the model's weaker performance on the
minority class.

Both forms of synthetic augmentation improved not-bombed recall,
balanced accuracy, and macro F1. The strongest results were obtained
with balanced DDPM augmentation, which achieved an accuracy of 0.88,
balanced accuracy of 0.81, macro F1 of 0.78, and not-bombed recall of
0.69. Relative to the real-only baseline, this corresponds to increases
of 0.14 in balanced accuracy, 0.13 in macro F1, and 0.28 in
not-bombed recall.

The balanced configurations outperformed the corresponding doubled
configurations for both generative approaches. This suggests that
targeted correction of the class imbalance was more effective than
proportionally increasing both classes while preserving their original
imbalance. DDPM augmentation also outperformed GAN augmentation under
both regimes.

\section{DISCUSSION}
\label{sec:discussion}

The results indicate that synthetic data are most useful when they
address a specific limitation of the real training set. Simply
increasing the number of samples while preserving the original class
imbalance produces smaller gains than targeted augmentation of the
underrepresented class.

DDPM augmentation performs better than GAN augmentation in both tested
regimes. One possible explanation is that iterative denoising provides
greater sample diversity and more stable coverage of the real feature
distribution. By contrast, GAN training on a small dataset is more
susceptible to mode collapse and repeated textures. This interpretation
should be verified using the feature-space precision, recall, diversity,
and nearest-neighbor analyses.

The results also show why overall accuracy is insufficient for this
task. The real-only model already achieves high bombed recall, but
performs poorly on not-bombed images. The largest practical benefit of
synthetic augmentation is therefore the improvement in balanced
accuracy, macro F1, and not-bombed recall rather than the smaller change
in overall accuracy.

\section{LIMITATIONS AND FUTURE WORK}
\label{sec:limitations}

The main limitation is the small size and restricted geographic scope
of the real dataset. All images originate from the same broader region
and satellite source, which limits conclusions about generalization to
different soil types, seasons, weather conditions, sensors, and regions
of Ukraine.

The binary task also provides a coarse representation of damage. A
patch is labeled bombed when at least one crater is present, regardless
of crater count, size, or damage severity. Future work should therefore
consider crater localization, instance counting, segmentation, and
object detection. For detection experiments, the main downstream
metrics should include mAP@0.5, mAP@0.5:0.95, precision, recall, and
class-specific average precision.

The generative models may reproduce training images or amplify
acquisition-specific biases. Exact duplicate checks are not sufficient
to exclude memorization, so nearest-neighbor and embedding-based
analysis should be included in the final evaluation.

Future experiments should also evaluate additional classifier
architectures, including ResNet and ConvNeXt, repeat all configurations
across multiple random seeds, and study several synthetic-data ratios.
External evaluation on imagery from additional regions and acquisition
periods would provide stronger evidence of generalization.

The proposed system is intended to support damage analysis and
prioritization. It cannot determine whether a field is safe or whether
unexploded damage is present.

\section{CONCLUSION}
\label{sec:conclusion}

This work investigated class-conditional GAN- and DDPM-based synthetic
augmentation for classifying battle-damaged agricultural fields in
Ukraine. Both generative models were trained only on real training
images, and all downstream evaluation was performed on a fixed real
test set.

Synthetic augmentation improved the class-balanced performance of the
Vision Transformer classifier. The strongest configuration used DDPM
samples to balance the training classes, increasing accuracy from 0.84
to 0.88, balanced accuracy from 0.67 to 0.81, macro F1 from 0.65 to
0.78, and not-bombed recall from 0.41 to 0.69. These results suggest
that targeted synthetic augmentation can be more effective than simply
increasing dataset size while preserving the original imbalance.

\bibliography{report}
\bibliographystyle{spiebib}

\end{document}